\pdfoutput=1
\documentclass[preprint,12pt]{elsarticle}

\usepackage{amssymb}
\usepackage{multirow}
\usepackage{subfigure}
\journal{IEEE Transactions}

\begin{document}

\begin{frontmatter}

\title{BGaitR-Net: Occluded Gait Sequence reconstruction with temporally constrained model for gait recognition}
\author[inst1]{Somnath Sendhil Kumar}
\author[inst1]{Pratik Chattopadhyay\corref{cor1}}
\affiliation[inst1]{organization={Pattern Recognition Laboratory, Department of Computer Science and Engineering},%Department and Organization
            addressline={Indian Institute of Technology (BHU)}, 
            city={Varanasi},
            postcode={221005}, 
            % state={Uttar Pradesh},
            country={India}}
            
\author[inst2]{Lipo Wang\corref{cor1}}

\affiliation[inst2]{organization={School of Electrical and Electronic Engineering},%Department and Organization
            addressline={Nanyang Technological University}, 
            city={Singapore},
            postcode={639798}}            

\begin{abstract}
Recent advancements in computational resources and Deep Learning methodologies has significantly benefited development of intelligent vision-based surveillance applications. Gait recognition in the presence of occlusion is one of the challenging research topics in this area, and the solutions proposed by researchers to date lack in robustness and also dependent of several unrealistic constraints, which limits their practical applicability.  
%such as tracking, gait recognition, re-identification, etc. Use of gait as a biometric to identify suspects in surveillance sites cannot be over-emphasized. However, to date its practical application is limited since the gait videos captured in surveillance sites are often corrupted due to occlusion which limits the effectiveness of traditional gait recognition techniques. In most of these previous attempts, gait-based person identification was done by extracting gait features from a complete gait cycle of the subject. This assumption is not suitable for real world scenarios with the subject being occluded in a few frames. There has been a few occlusion handling strategies for tackling this problem while recognizing the gait signatures, but these are rather limited and As a result, these cannot be potentially deployed % application %fail at reliably deploying in real-world scenarios. 
We improve the state-of-the-art by developing novel deep learning-based algorithms to identify the occluded frames in an input sequence and next reconstruct these occluded frames by exploiting the spatio-temporal information present in the gait sequence. The multi-stage pipeline adopted in this work consists of key pose mapping, occlusion detection and reconstruction, and finally gait recognition. While the key pose mapping and occlusion detection phases are done %using Constrained KMeans Clustering and 
via a graph sorting algorithm, %occlusion detection is carried out in this work by employing a custom built VGG-16 model. Following this, 
reconstruction of occluded frames is done by %computing an embedding of the images
%converting the images into vectors using Auto-Variational Encoders followed by 
by fusing the key pose-specific information derived in the previous step along with the spatio-temporal information contained in a gait sequence using a Bi-Directional Long Short Time Memory. This occlusion reconstruction model has been trained using synthetically occluded CASIA-B and OU-ISIR data, and the trained model is termed as Bidirectional Gait Reconstruction Network  \textit{BGait-R-Net}. Our LSTM-based model reconstructs occlusion and generates frames that are temporally consistent with the periodic pattern of a gait cycle, while simultaneously preserving the body structure, i.e., silhouette-level features at a high resolution. Finally,  GEINet-based classification is employed to identify the class of the subject from the reconstructed sequence.
% computed from the reconstructed gait cycle has been used as gait signatures to compare against a large set of gallery subjects. 
The %mainly novelty of the
effectiveness of our approach has been evaluated on the TUM-IITKGP and the synthetically occluded CASIA-B data sets, and encouraging results have been obtained.  %by recognizing the gait signatures of the reconstructed sequences. The Pose Energy is average of all images in a particular keypose set, henceforth we compute PEI for obtaining the gait recognition accuracy. Extensive evaluation of our approach is done on pubilc data sets and private datasets, and 
Comparative analysis with other popular state-of-the-art methods verify the suitability of our approach for potential application in real-life scenarios.

\end{abstract}


% \begin{highlights}
% \end{highlights}


\end{frontmatter}


\section{Introduction}
\label{sec:sample1}
Gait recognition refers to the process of identifying individuals from their gait signatures. Due to significant advancement in sensor quality and increased processing power for computer vision applications, high volumes of video data can be processed in a reasonable amount of time and useful conclusions can be inferred from the video data. This has motivated researchers to come up with plausible solutions to human identification from their walking videos captured by surveillance camera in public places such as railway stations, airports, and shopping malls. Over the past two decades, 
%This allows us to utilize gait-based identification for surveillance sites. 
there have been several attempts to develop suitable gait recognition algorithms for application in surveillance sites \cite{Chattopadhyay2015Oct} \cite{Babaee2019Apr} \cite{Roy2011Nov} where the gait videos are usually corrupted with occlusion. %but they depend on the availability of a complete gait cycle. Instead, in a practical scenario, the complete information may not be available due to the presence of occlusions. 
While the approaches discussed in \cite{Chattopadhyay2015Oct} \cite{Babaee2019Apr} attempt to perform recognition only from the available partial gait cycle information, other approaches like \cite{Roy2011Nov} uses a Gaussian Process Dynamical Model to predict the missing frames before performing recognition. Since gait is a periodic activity and the effectiveness of gait recognition depends on the availability of a complete cycle, the approach in \cite{Roy2011Nov} seems to be better than both \cite{Chattopadhyay2015Oct} and \cite{Babaee2019Apr} to carry out gait recognition in occluded situations. However, the assumption that walking pattern follows a Gaussian may not be true always. As expected, the accuracy of these approaches is not appreciably high and therefore deploying these methods in real-world scenarios will not produce any noticeable enhancement/improvement to the existing security setup in the above-mentioned surveillance sites. Moreover, the work in \cite{Roy2011Nov} reconstructs silhouettes forward in time, and does not utilize all the information present in the available frames. Presence of noise in a few frames is likely to degrade the quality of frame prediction, and the final gait cycle constructed after frame prediction using the above approach may not maintain the temporally consistency of a gait sequence.

%Occlusion %in gait cycles are majorly classified into can be broadly categorized as either static or dynamic occlusion \cite{Roy2011Nov}. %There have been attempts to use this partial information for gait recognition \cite{Chattopadhyay2015Oct}\cite{Babaee2019Apr}, but these do not produce good accuracy to use in a practical scenario. 
%Another approach is to reconstruct silhouette from the partial gait cycle and then recognize the gait. There have been a few attempts with this approach with a simple model-based approach that uses Gaussian Process Dynamic Model\cite{Roy2011Nov} and Deep learning-based reconstruction [2new]. These methods work on reconstructing silhouettes forward in time, which does not utilize all the information present and may not result in a temporally consistent cycle. 

In this work, we propose a deep bidirectional spatio-temporal model to reconstruct the occluded frames in the gait cycle. Unlike the approach in \cite{Roy2011Nov}, here we do not make any assumption regarding the distribution of the walking features over a gait cycle. Rather, we first identify occluded frames using a graph-sorting algorithm and next utilize the spatio-temporal information corresponding to all the unoccluded frames of a gait sequence to predict the missing/occluded frames. %obtain additional information about each silhouette, namely its key pose obtained from a modified K-Means Algorithm that inculcates temporal constraint followed by a graph-based sort \cite{Roy2012Mar}. 
%Prior to the occluded frame prediction, %a VGG-16 based network is used to detect the occluded frames present in the input sequence. Basically, this VGG-16 network takes as input each frame from a binary silhouette sequence and predicts whether occlusion is present in that frame or not. %It is very useful information in case of dynamic occlusions or silhouettes containing multiple subjects where we can replace the frame with a blank silhouette. 
The proposed occlusion reconstruction model has been termed as Bidirectional Gait Reconstruction Net (abbreviated as BGaitR-Net), and it is trained using the CASIA-B data \cite{Zheng2011Sep}\cite{Yu06aframework}, and the OU-ISIR Large Population Dataset \cite{Iwama2012Jun}. The effectiveness of our approach has been tested using the TUM-IITKGP \cite{Hofmann2011GaitRI} dataset which is the only dataset featuring both dynamic and static occlusions. However, this dataset consists of only a small set of walking sequences from 35 subjects. Hence, we choose two other datasets, namely  to train the proposed BGaitR-Net model.

The main contribution of our work may be summarized as follows:
\begin{itemize}
%\begin{itemize}
    \item We propose novel deep neural network architectures to reconstruct the occluded frames in a gait cycle in a temporally consistent manner from the spatio-temporal information present in the sequence and key poses in a gait cycle. 
    \item We demonstrate how synthetically occluded sequence data can be used along with the given unoccluded sequence as ground-truth to train a Bi-directional LSTM model to reconstruct the occluded frames present in any gait video sequence. To the best of our knowledge, applicability of learning-based time-series models have not yet been studied for the purpose of gait sequence reconstruction.
    
    %exploiting such spatio-temporal information present in the given gait sequence to predict occluded frames is a new approach, and it will also have wide applicability to advance the research in other Vision-related research domains such as tracking, re-identification, etc.
    \item Our approach has been evaluated through extensive experiments using the CASIA-B data corrupted with varying levels of synthetic occlusion and also using the real occluded sequences present in the TUM-IITKGP data. % that are previously done in literature. 
    The resources including the synthetically occluded data and pre-trained models have been made publicly available to the research community for further comparative studies. % among other methods and for future comparison.
%\end{itemize}
\end{itemize}
%and testing purposes, as these don’t have any occlusions (instead we introduced synthetic occlusions). The proposed occlusion reconstruction model has been termed as Bidirectional Gait Reconstruction Net (abbreviated as BGaitR-Net).

\section{Related Work}
\label{sec:sample2}
%\subsection{Overview of Research on Gait Recognition}
%\label{subsec:sample1}
Traditional gait recognition approaches can be classified as either appearance-based or model-based. While the appearance-based approaches extract gait features from the silhouette shape variation over a gait cycle, the model-based methods attempt to fit the kinematics of human motion in a pre-defined walking model. Appearance-based approaches have become more popular over the years due to their ease of implementation and less computational requirements. Among this category of methods, the work in \cite{Han2006Feb} presents a feature called the Gait Energy Image (GEI) that computes the average of gait features over a complete gait cycle. Due to aggregating features over a gait cycle, the GEI cannot capture the dynamics of gait effectively. Later on, a few approaches have been developed that have made attempts to solve the limitations of GEI. As an example, the work in \cite{Roy2012Mar} introduces a pose-based feature by aggregating features from fractional parts of a gait cycle. This feature is termed as the Pose Energy Image (PEI) and it has the potential to capture the kinematics of gait at a higher resolution. A few similar fractional gait cycle-based feature extraction techniques can be seen in \cite{Chattopadhyay2014Jan} and \cite{Chattopadhyay2015Oct}] that uses the RGB, depth, and skeleton streams from Kinect. Another improvement over the GEI is described in \cite{Zhang2010Jul} in which the active walking regions are computed by subtracting adjacent binary silhouette frames, and next these difference images are aggregated to compute the Active Energy Image (AEI) feature. %Gait recognition is done by projecting the AEI features into the LPP subspace. 
In \cite{Xu2007Oct}, the GEI features are first projected into a lower-dimensional space suing Marginal Fisher Analysis, and recognition is done using the sub-space features. A view point invariant gait recognition approach described in \cite{Collins2002May} performs cyclic gait analysis to identify the key frames present in a walking sequence. Standard structural features such a height, width, different body-part proportions, stride length, etc., have been used for recognition via normalized correlation. All the above-mentioned approaches require a complete cycle of gait for proper functioning and hence, are not suitable for performing gait recognition in presence of occlusion. % occluded sequence are provided.

%While the appearance-based approaches extract gait features from silhouette shape variation over a gait cycle, the model-based methods attempt to fit the kinematics of human motion in a pre-defined. A few methods developed attempt to fit the kinematics of human motion in a pre-defined. 
A few popular model-based gait recognition methods are discussed next. In \cite{Ariyanto2011Oct}, an articulated 3D points cylinder has been considered as a gait model. A 3D voxel representation of each walking stance is computed from multiple cameras which is fitted to the above 3D model. Static gait features are extracted from the height, stride length, and other body-part dimensions of the fitted model, whereas the kinematics of gait are extracted from the lower part of the body, more specifically, by encoding the variation of thigh and shin angles over a gait cycle. In \cite{Cunado2005Jun} also, gait features are extracted from the lower part of the body. Hough transform is applied to each frame of a binary silhouette sequence to obtain the boundaries of the legs of the silhouette in that frame. The variation in the inclination of these lines is next fitted to a simple harmonic motion model from which the gait features are computed. The work in \cite{Bobick2001Dec} considers a skeleton model and fits static body structural parameters such as distance from neck to torso, stride length, etc., to the model. First body part labelling is done from the silhouette images, following which depth compensation is done to convert pixel information into 3D world coordinates, which is henceforth used to compute the static body parameters. In \cite{Cunado2003Apr}, another model-based approach is discussed which also predicts the body parts and extracts features from the Fourier Transform of the motion data derived from these body parts. Finally, a K-NN based classification method is adopted to predict the class of a subject. In \cite{Lee2002May}, the centroid of each silhouette is computed based on which the silhouette is divided into seven segments. An ellipse is next fitted to the foreground part of each segment and a set of static parameters of the ellipse such as the centroid, ratio of the major to minor axis length, etc., are considered for computing the gait features. %Model-based approaches are generally invariant to viewpoint changes, but high computational overhead and requirement of high-resolution images limit their practical applicability in real-life scenarios. Also, computation of appearance-based gait features can be done quite accurately without undergoing the complex model fitting operations, and hence appearance-based methods gained higher popularity compared to the model-based approaches. Early approaches on appearance-based gait recognition mostly use gait videos captured by RGB cameras and focus on the fronto-parallel, i.e., side view of gait, since binary silhouettes from the fronto-parallel view provide maximum gait information. In the later years, a few cross-view gait recognition were propesed in which the training and the test sets are captured from different views. For example, the work in \cite{Ben2019Jan} performs recognition by aligning the GEIs\cite{Han2006Feb} from different viewpoints using coupled bi-linear discriminant projection (CBDP).

With the advancement of Deep Learning, CNN-based models have also been studied for gait recognition. For example, in \cite{Hu2018} and \cite{YantaoLiJul2020}, raw sensor data from the accelerometer and gyroscope of smartphones are used to monitor users' behavioral patterns. A CNN architecture is trained using the temporal and frequency domain data to extract an information rich feature representation. Next, SVM-based classification of these features is done in the latent space to map a person as either a legitimate user or an imposter. In \cite{Qin2021}, a single shot learning-based palm vein identification technique is proposed in which at an initial step data augmentation is done by passing a single image from each class through a series of GANs to generate multiple instances of that class, and these augmented images are next used to train a CNN. recently, CNNs have also been used for cross-view gait recognition, for example, the work in \cite{Takemura2017Oct} describes a deep Siamese architecture-based feature comparison that works satisfactorily even for large variation of view angles. Among the other recent deep learning-based gait recognition approaches, in \cite{Shiraga2016Jun} the GEI features computed from a gait cycle are passed through a CNN-based model, termed as \textit{GEINet}, to obtain deep features which are next used for classification. Since training data in the case of gait recognition is scarcely available, training a deep network may suffer from under-fitting. Hence, instead of using a deep CNN model, typically consisting of a large number of trainable parameters, the authors in \cite{Alotaibi2017Nov} suggest employing a small-scale CNN consisting of four convolutional layers (with eight features maps in each layer) and four pooling layers for gait recognition.

With the introduction of RGB-D cameras such as Kinect, a few frontal-view gait recognition techniques \cite{Chattopadhyay2014Jan, Sivapalan2011Oct} have also been developed. An advantage of frontal view gait recognition is that it is less prone to occlusion, as a result of which there is a higher chance of capturing clean and usable gait cycle information even from a short sequence. Since, reliable gait features cannot be extracted from frontal view binary silhouette sequences, depth streams provided by depth cameras such as Kinect have been mostly utilized in research on frontal gait recognition. The work in \cite{Battistone2019Sep} jointly exploits structured data and temporal information using a spatio-temporal neural network model, termed as the TGLSTM, to effectively learn long and short-term dependencies along with a graph structure. Initially, a graph is constructed from each frame containing a binary silhouette that represents the skeleton structure of the silhouette in the frame. Following this, an LSTM is used to capture the variation of the skeletal joint features over consecutive frames. However, the effectiveness of this method is likely to suffer if any input silhouette frame is corrupted by noise. 

The gait recognition scenarios considered by the above techniques are very simplistic in the sense that these consider only one person to be present in the field of view of a camera, However, an ideal gait recognition algorithm must not be dependent on such constraints. Handling occlusion in gait recognition effectively is extremely essential since occlusion is an inevitable occurrence in any practical situation. The presence of occlusion makes the silhouettes in the video frames noisy and often hinders the capturing of a complete clean gait cycle. This affects the recognition accuracy of most traditional appearance-based approaches, as discussed before. Some popular approaches towards handling the problem of occlusion in gait recognition are discussed next. Occlusion reconstruction has been done using a Gaussian process dynamic model in \cite{Roy2011Nov}. In this work, occluded frames in a gait sequence are  first detected and next these occluded frames are reconstructed by assuming that the variation of gait features over a gait sequence can be modelled by fitting a multi-variate Gaussian curve. The viability of this approach has been evaluated using the TUM-IITKGP data \cite{Hofmann2011GaitRI}. In \cite{Isa2010Dec}, the authors proposed an approach that uses Support Vector Machine-based regression to reconstruct the occluded data. This reconstructed data is first projected onto the PCA subspace and next the projected features are classified to the appropriate class %has been carried out 
in this canonical subspace. Three different techniques for the reconstruction of missing frames have been discussed in \cite{Lee2009Dec}, out of which the first approach uses an interpolation of polynomials, the second one uses auto-regressive prediction, and the last one uses a method involving projection onto a convex set. In \cite{Aly2014Dec} an algorithm focusing on tracking pedestrians has been presented in which the results are evaluated on a synthetic data set containing sequences with partial occlusion. Some approaches involving human tracking \cite{Zhang2010Aug, Andriluka2008Jun} and activity recognition \cite{deLeon2002Aug, Weinland2010Sep} techniques have also been developed which handle the challenging problem of occlusion detection and reconstruction.

From the extensive literature survey it can be seen that gait recognition in the presence of occlusion is still an open area of research and effectiveness of Deep Neural Network-based models to predict the missing/occluded frames has not been studied yet. In this work, we specifically focus on this aspect and propose a new VGG-based model to detect occluded frames in a gait sequence and a spatio-temporal LSTM-based architecture to reconstruct these occluded frames. The proposed approach is explained in further detail in the following section. 

\section{Proposed approach}
\begin{figure}[ht]
\centerline{\includegraphics[width = \textwidth]{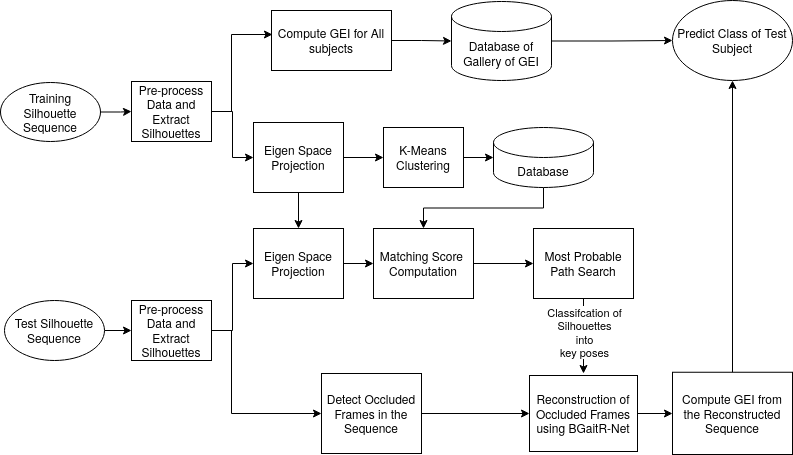}}
\caption{A flowchart of the pipeline of the proposed algorithm}
\label{fig:proposed_approach}
\end{figure}

A schematic diagram explaining the steps of the proposed approach, namely the occlusion reconstruction module and key pose identification \cite{Chattopadhyay2014Jan} with occlusion detection is shown in  Figure~\ref{fig:proposed_approach}. Throughout the text, we assume that the gallery sequences are always unoccluded, and the test sequences are corrupted due to occlusion. 
With reference to the figure, standard pre-processing steps \cite{Han2006Feb,Zhang2010Jul,Shiraga2016Jun} are first applied to extract the binary silhouettes from the RGB frames and normalize these. This step involves background subtraction using a suitable technique, cropping out the region of interest and resizing each cropped region to a fixed height and width. The objective is to obtain binary image frames of pre-defined dimensions, with each frame containing a silhouette such that the silhouette region is colored white and the background is colored black (refer to Fig. \ref{rec_img}). The pre-processing steps are already explained in detail in the above cited papers, and hence we are not extending discussion on this topic further. Once the pre-processed silhouette sequences are obtained from all the gallery sequences, we compute a database of gallery GEI features \cite{Han2006Feb} for each subject, and also a database of a generic set of key poses in the PCA-subspace from all the subjects as explained in \cite{Roy2011Nov,Roy2012Mar,Chattopadhyay2014Jan}. 
%The key pose identification is a temporally constrained K-Means Clustering algorithm \cite{Chattopadhyay2014Jan} 
These key poses are representative poses of a gait cycle are constructed in a suitable manner so as to preserve the temporal order of walking in a gait cycle. The information about the generic key pose set is used during the testing phase for effective frame reconstruction in an occluded sequence. Now given an occluded test sequence, similar pre-processing steps are followed, and next a graph-based algorithm is followed to determine the probable key pose to which each frame of the sequence can be mapped to. Next, reconstruction of the occluded frame is done by employing the proposed BGaitR-Net using the encoding vector corresponding to the probable key pose as an auxiliary information. Finally, gait recognition is done using the the reconstructed sequence to identify the class of the test subject. %that segments the sequence into a fixed number of key poses and gives the key pose for each image, this information is used as an embedding with the encoded vector of the image for better reconstruction. The figure explains the process of person identification of a test subject being done by comparing his/her gait signatures against a gallery of a large number of subjects. However, in a real-life scenario, occluded frames are generally not annotated hence we used an automated mechanism for occlusion detection using a VGG-16 based deep model. So after the detection of occluded frames, they are reconstructed in a window of 6 frames. This reconstruction is similar to filtering the Spatio-temporal series. Hence the algorithm considers the images before the occluded frame and also the images after the occluded frame. This reconstruction is done using a Bi-LSTM based model that takes the window of six frames and returns the reconstructed 6 frames. Finally, the Gait Energy Image (GEI) is computed from this sequence, Then the GEI of the test subjects are compared against the GEI features of a large gallery of subjects using a Convolution-based Neural Network. % \cite{Lecun1998Nov}. 
The individual steps of the proposed approach are explained in detail next.

\subsection{Identifying Occluded Silhouettes and Key Pose for Non-Occluded Silhouettes} \label{subsec:vgg}
\begin{figure}[ht]
\centerline{\includegraphics[width = \textwidth]{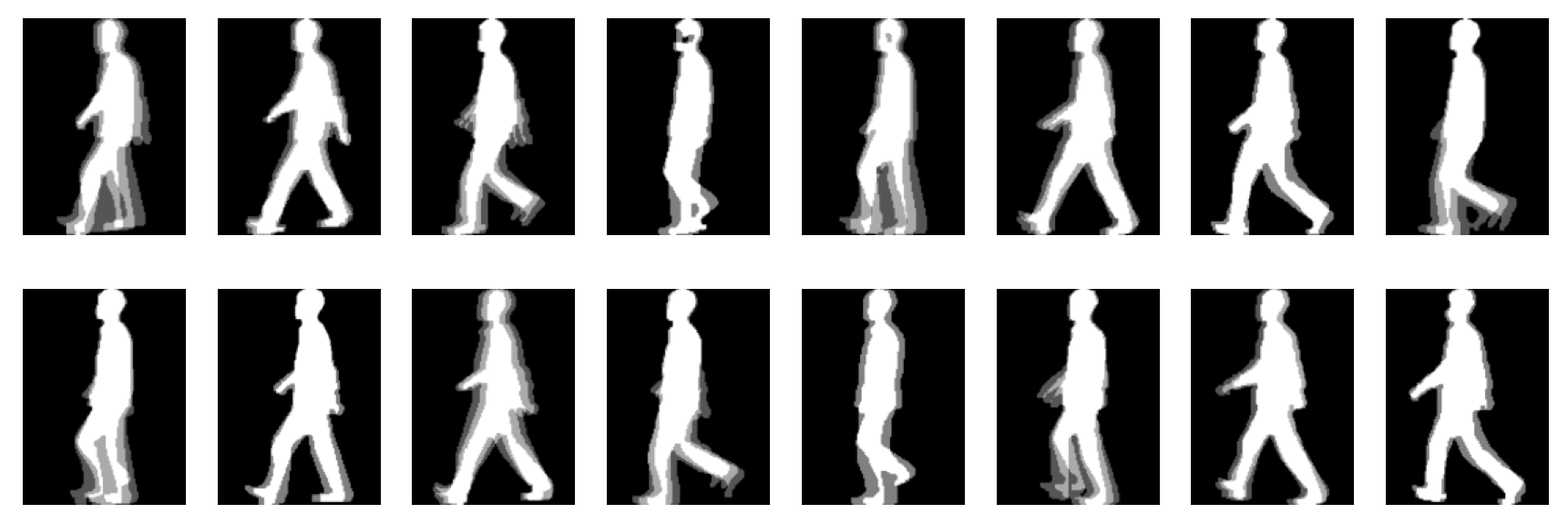}}
%\centerline{\includegraphics[width= \textwidth]{enc_Dec.png}}
\caption{A set of 16 key poses}
\label{fig:Keyposes}
\end{figure}

%%%%%%%%%%%%%%%%%%%%%%%%%%%%%%%%%%%%%%%%
%The set of generic key poses is next used to find a mapping between 
As discussed before, the occlusion reconstruction model requires auxiliary information regarding the pose of the subject in each frame of the silhouette sequence. The pose information must be fed to the model as a vector embedding. For this, we first need to determine the vector embedding corresponding to the key poses (or, key walking stances) present in any gait cycle \cite{Roy2012Mar,Chattopadhyay2014Jan}. The key poses are generic and are not specific to a particular individual. We construct the key poses by applying constrained K-Means clustering on a large number of aligned sequences in a manner similar to that described in  \cite{Roy2012Mar,Chattopadhyay2014Jan}. Since the same key pose construction method has been extensively used in several studies on gait recognition in the past, we have not included this algorithm in the present paper. We use several gait cycles from the CASIA B and the OU-ISIR data to obtain the key pose embedding in the PCA-subspace. The decoded images corresponding to the key pose embedding are shown in Fig. \ref{fig:Keyposes}.

\begin{figure}[ht]
\centering
\includegraphics[scale = 0.3]{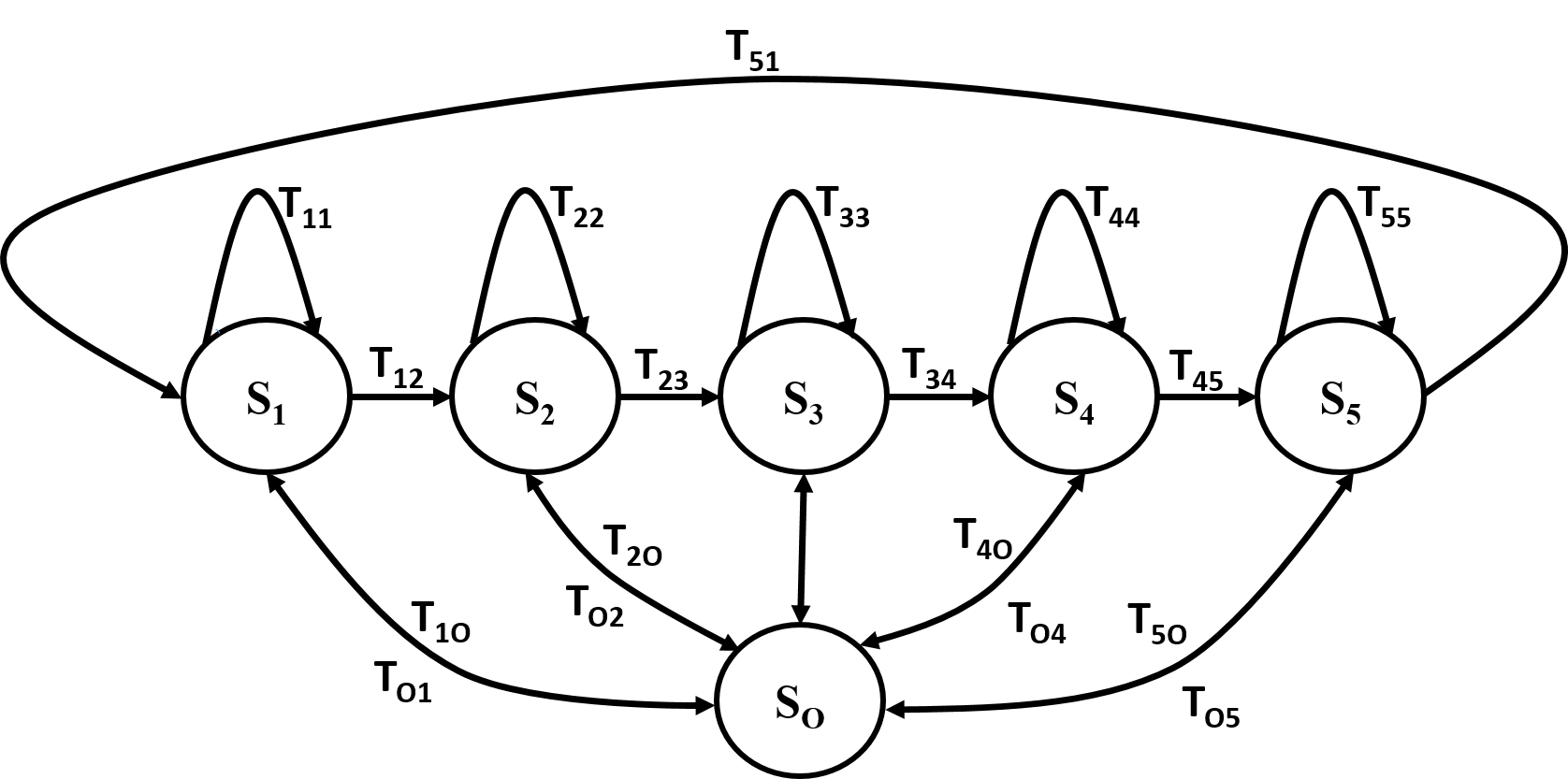}
\caption{A state transition model to map frames to the appropriate key poses}
\label{stm}
\end{figure}

The silhouettes corresponding to the frames of an input gait sequence are next classified into the appropriate key poses. It maybe noted that direct mapping of frames to key poses based on Euclidean distance (or any other distance metric) may be erroneous since this process does not consider the temporal relation between the adjacent frames in a sequence, and two consecutive frames may get mapped into non-consecutive key poses. A better technique would be to carry out frame classification using a state transition model as also done in \cite{Roy2011Nov}. A representative state transition model with five states is shown in Fig. \ref{stm}. 
In this model, each key pose is represented as a key pose state, and the transition between these states are shown with arrows. Ideally, for $K$ key poses, the state transition model should have $K$ key pose states ($S_1$, $S_2$, ... $S_K$), and one occlusion State ($S_O$), resulting in total ($K$+1) states.  However, in the figure only 5 key pose states, namely, $S_1$, $S_2$, $S_3$, $S_4$, and $S_5$ and one occlusion state $S_O$ are shown for ease of representation and explanation. A transition from state $S_i$ to $S_j$ is represented as $T_{ij}$. From the figure, it can be seen that from a given state, transition is allowed to the same state or to the immediately subsequent state in a cyclic manner. As an example, from $S_3$, transition is allowed either to $S_3$ or to the next state $S_4$, whereas from the final state $S_K$, transition is allowed either to $S_K$ or to $S_1$. Since occlusion can occur anywhere in a walking sequence, transitions may be possible from any of the $K$ states to $S_O$, and also from $S_O$ to any of the $K$ key pose states. Since the occluded state $S_O$ is also the ($K$+1)$^{th}$ state as per Fig, \ref{stm}, in further discussions by $S_{k+1}$ we will mean $S_O$.

The problem of silhouette classification into the appropriate key poses using the state transition diagram is next mapped to the problem of finding the shortest path in a Directed Acyclic Graph (DAG). We construct a vertex-weighted DAG from the above state transition model. Let, the input binary silhouette sequence of frames be denoted by \textbf{F} = $F_1$, $F_2$,..., $F_N$. First, Euclidean distance is computed between the encoded representations of each pair $F_i$ and $S_k$,  \textit{i}=1,2,...,\textit{N}, and \textit{k} = 1,2,...,\textit{K}+1, and a matrix $M$ is constructed from these distance measures. The dimensions of this matrix is $N{\times}(K+1)$, and each cell of this matrix corresponds to the Euclidean distance between the $i^{th}$ frame and the $k^{th}$ state, i.e., $M_{ik}$ = $D(F_i,S_k)$, where $D$ denotes the Euclidean distance.

\begin{figure}[ht]
\centerline{\includegraphics[width = \textwidth]{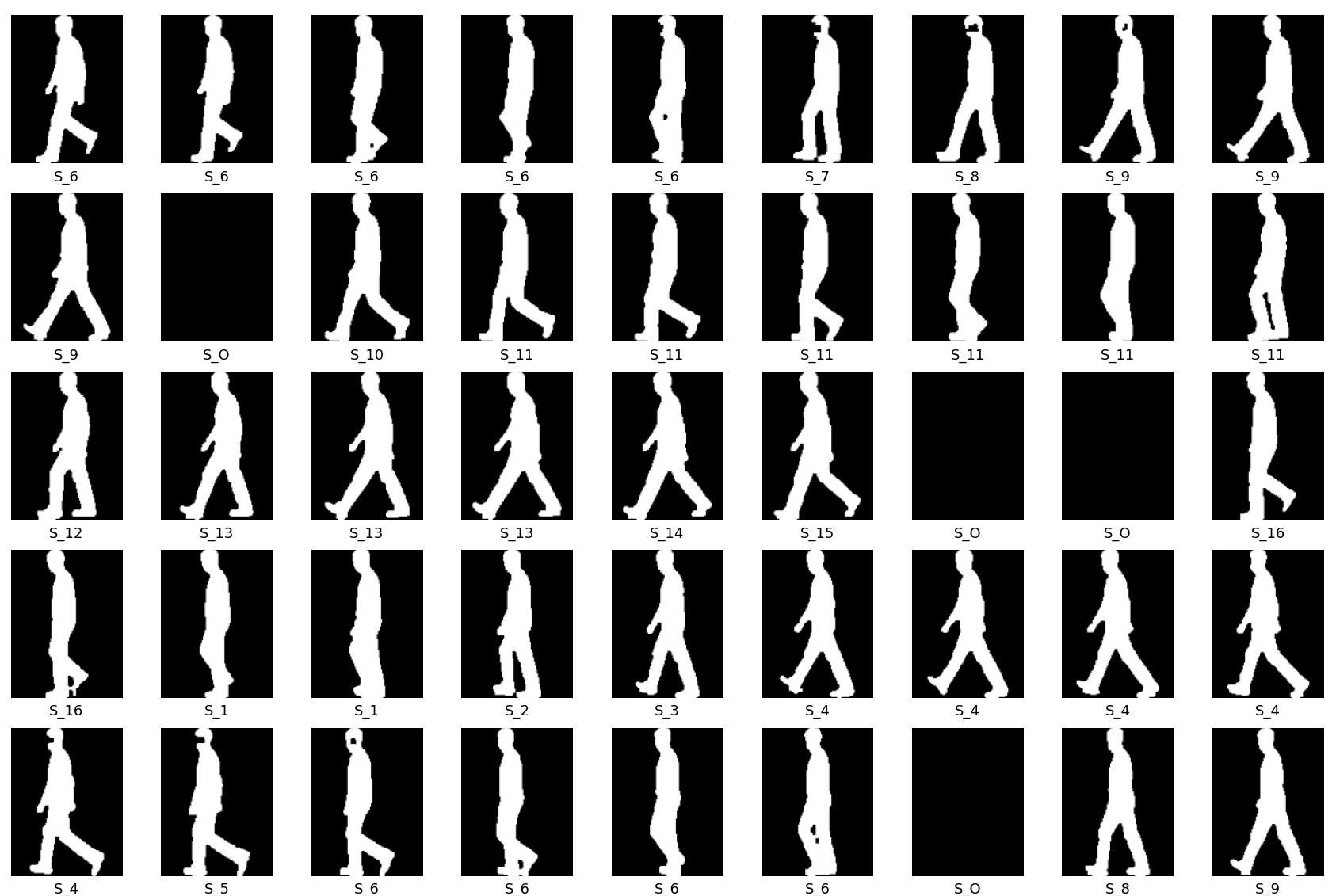}}
%\centerline{\includegraphics[width= \textwidth]{enc_Dec.png}}
\caption{An occluded frame sequence and the mapped states corresponding to each frame}
\label{fig:occ_kmean}
\end{figure}

In the DAG, there are $N(K+1)$ vertices corresponding to the 
$N(K+1)$ cells of the $M$ matrix. If we use the pair \{$<frame\_no>,<state\_no.>$\} to represent each vertex in the DAG, then an edge is added directed from \{$F_i$,$S_{k_1}$\} to \{$F_{i+1}$,$S_{k_2}$\} only if there exists an allowable transition from state $S_{k_1}$ to state $S_{k_2}$ (1$\leq$ $k_1$,$k_2$ $\leq$ $K$+1), as per the state transition model in Fig. \ref{stm}. The weight associated with vertex \{$F_{i+1}$,$S_{k_2}$\} is $M_{ik}$, and the edges are not weighted. Finally, we find the path in the graph along which the sum of the vertex weights is minimum. Dynamic programming-based shortest path finding algorithm has been employed to find the most probable path, and the algorithm followed here is similar to the one used in \cite{Roy2011Nov}. Fig. \ref{fig:occ_kmean} shows an input synthetically occluded binary silhouette sequence with 45 frames and the corresponding states to which each frame gets mapped to using the above algorithm. \textbf{We have to write how we are representing the occluded state.} \textit{The state Here is not important the one hot encoding part has been stated while describing the model input for the encoder}

%%Figure on an input frame sequence and corresponding key pose state prediction

\begin{figure}[ht]
\centerline{\includegraphics[width = \textwidth]{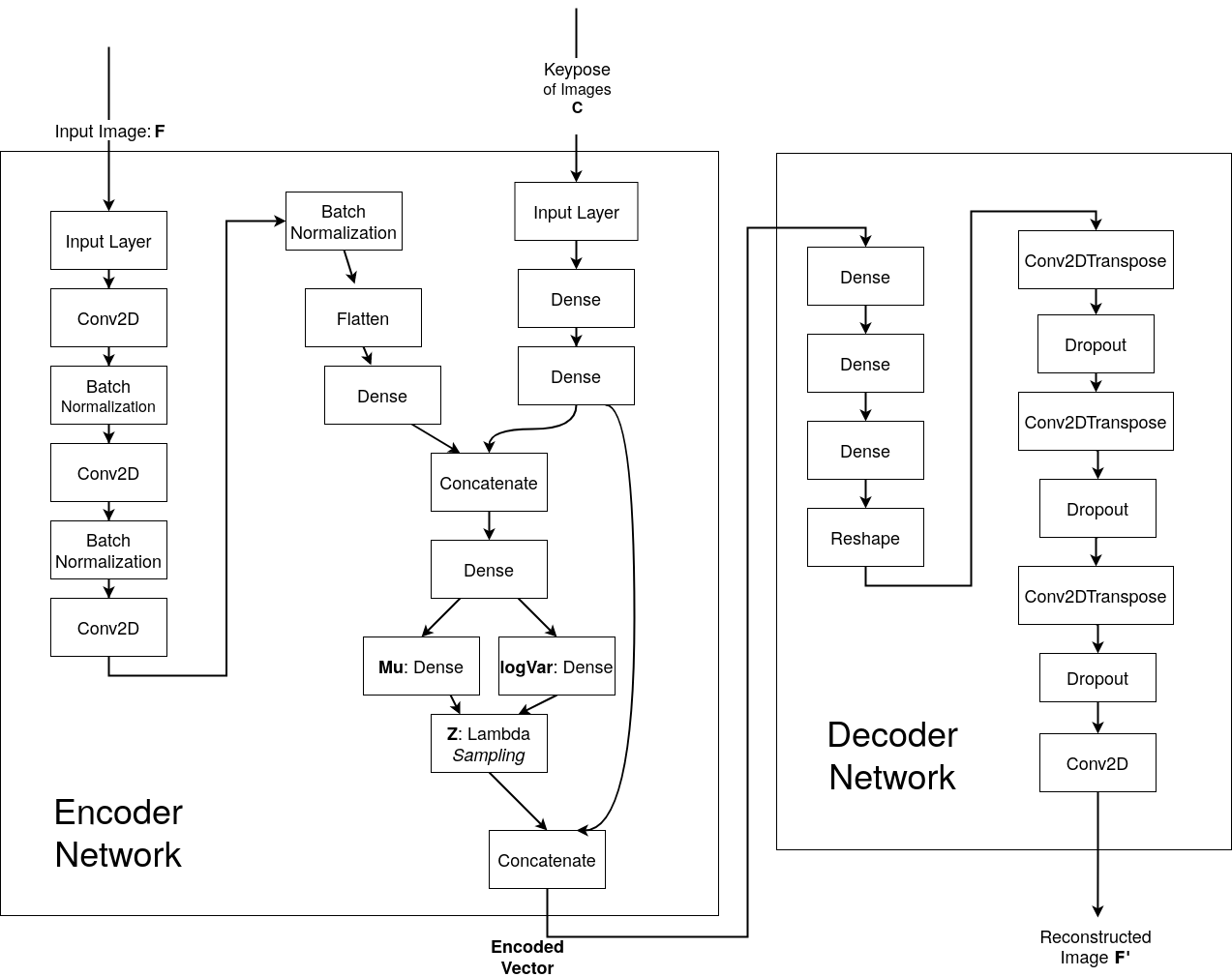}}
%\centerline{\includegraphics[width= \textwidth]{enc_Dec.png}}
\caption{Model architecture of the Variational AutoEncoder}
\label{fig:enc_dec_model}
\end{figure}

\begin{figure}[ht]
\centerline{\includegraphics[scale=0.45]{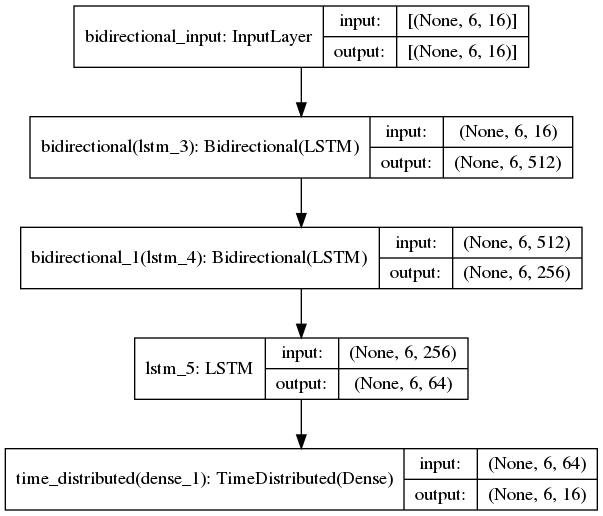}}
\caption{Model architecture of the Bi-LSTM for reconstruction}
\label{fig:bilstm_reconst}
\end{figure}

\subsection{Occlusion Reconstruction using BGaitR-Net}\label{occlusionReconstruction}
%After computing the key poses \cite{Roy2012Mar} of all the images in the sequence we employ the VGG-16 based occlusion detector Section \ref{subsec:vgg} from which we mask these occluded images with blank images to achieve better performance in reconstruction. 
The frames that have been identified as occluded by the above state transition model are only used for reconstruction by the proposed \textit{BGaitR-Net}. Any binary silhouette sequence corresponding to human walking can be viewed as a spatio-temporal data in which silhouette images form a periodic progression and hence these silhouettes can be said to be related to each other across the temporal domain. It appears that the binary silhouette frame at a given time can be predicted by exploiting the spatio-temporal information present in the gait sequence along with the %of each image within a given frame of time can be predicted and filtered with information from frames them self along with the 
additional information about the probable key pose corresponding to the frame. 
%including the dynamics of the human subject achieving better prediction. This way the occluded/corrupted frames are reconstructed and be used further in filtering. 

In this work, we propose to convert the images into a latent vector and then filter a window of frames using a deep time-series-to-time-series neural network, more specifically a Bi-Directional Long-Short Term Memory(Bi-LSTM) network \cite{Schuster1997Dec}. In the past BiLSTMs have been successfully used for time series filtering. To encode the silhouette into latent vectors we use Conditional Variational Autoencoder \cite{Pagnoni2018Dec}, which is similar to traditional Variational AutoEncoder\cite{Kingma2013} with a conditional vector corresponding to the embedding of the key pose of the silhouette. Since the gait of a person is in temporal progression, its latent vector can be filtered with the Bi-LSTM, and hence reconstructing the filtered latent vector with the decoder network will provide us the desired reconstructed gait sequence. The Conditional Variational Auto-Encoder used here consists of three convolutional layers with three Batch Normalization Operations and four Dense layers which estimate the parameters for the normal probability distribution of the images latent vector which is sampled and then used as the encoded latent vector. We are using three fully connected layers to encode this key pose information into the conditional vector. This key pose information is a One Hot encoding, where the starting 16 positions in the vector corresponds to the 16 Keyposes considered for the experiment. The last bit represents if the frame is occluded or not, i.e., If the frame is occluded we assign the last bit as 1 else we assign the last bit as 0 and assign 1 to the index mapped to the frames keypose.

The Encoder network learns a function $E$ that takes a binary silhouette frame($F$) from the sequence, and the conditional key-pose vector $c$ (i.e., key pose embedding) as input to generate parameters of a normal distribution i.e., $\mu$ and $log(\sigma)$, these denote the mean of a normal distribution and the log of the variance of a normal distribution respectively. Using these parameters we get a probability distribution from which we sample a latent vector represented by $z$. This can be mathematically represented by the Eqns. (\ref{encoder_model}) and (\ref{normal_sampling}). 
\begin{equation} \label{encoder_model}
     [\mu, log(\sigma)] = E(F,c)\, ,
\end{equation}
\begin{equation} \label{normal_sampling}
     z \sim \mathcal{N}(\mu,\, \sigma^{2})\, ,
\end{equation}
Let the Decoder network learn a function that takes the input of latent vector $z$ and the condition vector $c$ to return the reconstructed Frame $\hat{F}$. Then, $\hat{F}$ can be mathematically represented by Eqn. \ref{decoder_model}:
\begin{equation} \label{decoder_model}
    \hat{F} = D(z, c)\, .
\end{equation}
The Encoder-Decoder network has been trained using the Reconstruction loss ($L_{rec}$) %for the Output Image of the Encoder-Decoder network given by 
as shown in Eqn. (\ref{rec_loss}), which is defined as the binary cross-entropy loss between the input silhouette and the reconstructed silhouette. A second loss function used in to train the model is the Kullback-Leibler divergence ($L_{kl}$) loss for the normal probability distribution of the latent vector of the images is given by Eqn. (\ref{kl_loss}). 
\begin{equation} \label{rec_loss}
    L_{rec} = - \frac{1}{WH}\sum_{i=0}^{W} \sum_{j=0}^{H} (F(i,j)log(\hat{F}(i,j)) + (1-F(i,j))log(1-\hat{F}(i,j)))\, ,
\end{equation}
%\begin{flushright}
where $W$ and $H$ are the width and height of the input silhouette.
%\end{flushright}

\begin{equation} \label{kl_loss}
    L_{kl} = \sum\, (\mu^{2} + \sigma - log(\sigma) - 1) \, ,
\end{equation}
Incorporation of $L_k$ ensures compact and meaningful encoding of the images into the latent vector. Hence the loss function for the Conditional Variational Auto-Encoder ($L_{cvae}$) is the weighted summation of the two losses and is given by Eqn. (\ref{cvae_loss}), where Where $\lambda_1$ and $\lambda_2$ are two user-defined constant parameters. 
\begin{equation} \label{cvae_loss}
    L_{cvae} = \lambda_1L_{rec} + \lambda_2L_{kl} \, .
\end{equation}
In this work, the value of $\lambda_1$ and $\lambda_2$ are set to \textit{1} and \textit{0.5}, respectively.

The Bidirectional LSTM architecture used in this work is schematically shown in Figure \ref{fig:bilstm_reconst}. This network consists of 3 bidirectional Time-distributed layers and one Time-distributed LSTM network. This takes in 6 latent vectors represented by $Z$ (\ref{col_z}) as input and returns 6 reconstructed latent vectors $\hat{Z}$ as output. Let us represent the function learned from the network as $T$ then it can be mathematically represented by \ref{col_z_pred}:

\begin{equation} \label{col_z}
     Z = \{ z_{1}, z_{2}, z_{3}, z_{4}, z_{5}, z_{6} \}
\end{equation}

\begin{equation} \label{col_z_pred}
    \hat{Z} \equiv \{ \hat{z_{1}}, \hat{z_{2}}, \hat{z_{3}}, \hat{z_{4}}, \hat{z_{5}}, \hat{z_{6}} \} = T(Z)
\end{equation}

The Bi-LSTM model was trained using synthetic occlusions introduced using as Random Variate to decide to occlude a particular frame or not. This allows us to train the Bi-LSTM with varying degrees of occlusions. This can be represented mathematically using \ref{z_masked} where $Z$ is the latent vector of 6 consecutive frames, $X$ is the random variate of a Uniform distribution, $deg_{occ}$ is the degree of occlusion required in the dataset and the resulting dataset is $Z_{occ}$.

\begin{equation} \label{x_col}
    X = \{ x_1, x_2, x_3, x_4, x_5, x_6 \}\, , x_i \sim \mathcal{U}(0,1) \, , 
\end{equation}

\begin{equation} \label{z_masked}
    Z_{occ} = (X > deg_{occ})*Z \, ,
\end{equation}

\begin{equation} \label{rec_pred}
    \hat{Z} = T(Z_{occ})\, .
\end{equation}

The model is trained using Mean Squared Error loss($L_{mse}$) of the Original Latent vectors($Z$) and the predicted latent vectors($Z_{oc}$). This is mathematically given by \ref{mse_loss}.

\begin{equation} \label{mse_loss}
    L_{mse} = \frac{1}{n}\sum^{n}\, \sum_{i=1}^{6} (z_i - \hat{z_i})^2 \, .
\end{equation}

\subsection{Gait Recognition Using GEINet}
The effectiveness of the reconstruction model has been tested by carrying out gait recognition using the \emph{GEINet} model \cite{shiraga2016geinet}. This is a CNN-based model which is trained using the GEIs of all the gallery subjects, and the class of a test subject is predicted by inputting the GEI of the test subject. The same architecture of the \textit{GEINet} as given in \cite{shiraga2016geinet} has also been used in this work. The model has two convolutional layers with 18 and 45 kernels, respectively with max-pooling and ReLU activation in each layer. Let us consider a total of $\mathcal{S}$ sequences in the gallery and for each sequence (say, $\mathcal{S}_i$) we have the class label ($\mathcal{C}_i$) information and we also compute the corresponding GEI (say ${GEI}_i$), (\textit{i} = 1,2,...,$\mathcal{S}$). The training set used to train the \textit{GEINet} is formed from pairs of GEI and the class label as follows:
\{($GEI_1$,$\mathcal{C}_1$),($GEI_1$,$\mathcal{C}_2$),...,($GEI_\mathcal{S}$,$\mathcal{C}_\mathcal{S}$)\}. 

\textbf{The model is trained for 50 epochs at which point it was seen to achieve convergence and the average validation accuracy reached the 98\% mark and did not show any significant improvement with further training.}

\section{Experimental Setup}
The proposed algorithm has been trained on a system with 192 GB of RAM and 16 Xeon(R) CPU E5-2609 @ 1.7 GHz and 7 GeForce GTX 1080 Ti with 11 GB RAM, 11 GB frame-buﬀer memory and 256 MB of BAR1 memory and one Titan XP with 12 GB RAM, 12 GB frame-buﬀer memory and 256 MB BAR1 memory. Testing of the algorithm was done on a modest system with 16 GB of RAM and 1 Ryzen 5 3550H at 2.1GHz and GeForce GTX 1650 Ti with 4 GB RAM, 4 GB frame-buffer memory, and 128 MB of BAR1 memory. Hence, all test performance results of the algorithm provided are from the second Ryzen-based system. This is done to emphasize the usability of our proposed algorithm in a constrained setup. The OU-ISIR large population data set \cite{Iwama2012Jun} consists of walking sequences from more than 3000 subjects, and has been used for training the BGaitR-Net model only in combination with the CASIA-B \cite{Yu06aframework}. Evaluation of the proposed algorithm has been done on two public data sets, namely the TUM-IITKGP \cite{Hofmann2011GaitRI} which consists of occluded sequences and the CASIA-B \cite{Yu06aframework} which contains only unoccluded sequences but has been corrupted by adding varying levels of synthetic occlusion. 
%The main reason not to use the OU-ISIR large population data set \cite{Iwama2012Jun} is that size of the gallery is large with over 3000 subjects so we only used it for training the AutoEncoder and did not use it for gait recognition. Among the two datasets used, 

\subsection{Dataset for gait reconstruction and gait recognition}\label{dataset}

The CASIA-B dataset consists of 124 subjects and for each subject there are 6 different sequences which can be divided into 3 sets: (a) six sequences with normal walking (nm-01 to nm-06), (b) two sequences with carrying bag (bg-01 and bg-02), (c) two sequences with wearing a coat (cl-01 and cl-02). For gait recognition, we use normal walking sequences (i.e., sequences nm-01 to nm-06). Four of these sequences are used for training and the remaining two are used for testing our gait recognition model. The TUM-IITKGP dataset, on the other hand, consists of walking videos of 35 subjects under the following scenarios: (a) one video of normal walking without any occlusion, (b) one with carrying bag, (c) one with wearing gown, (d) one with static occlusion, and (e) one with dynamic occlusion. In the present evaluation, we have used normal walking to train the recognition model and construct the GEI gallery for all 35 subjects. For testing the algorithm we have used videos with static and dynamic occlusions. We have used 22.85\% of subjects from the gallery of 35 subjects at a time for gait recognition just as proof of concept to show the efficiency on our gait reconstruction method for recognizing humans from their gait signatures. For the evaluation of the proposed algorithm, we have selected occluded sequences for the corresponding subject. Throughout this section, the aforementioned test sequences of the TUM-IITKGP dataset are labeled as Sequence1, Sequence2, ...,  Sequence8 respectively.

\subsection{Qualitative and Quantitative analysis of the result of the gait reconstruction}\label{quaGaitrecon}

\begin{figure}[ht]
\begin{subfigure}[53\% occlusion reconstruction]{
\centering{\includegraphics[width = \textwidth]{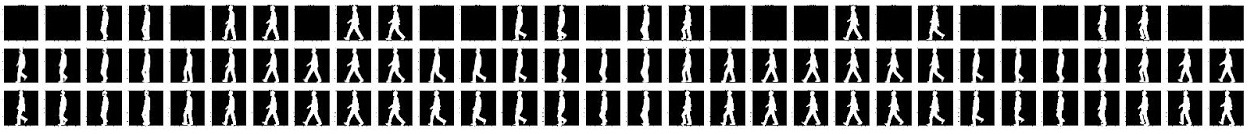}}
\label{rec_img:a}}
%\caption{53\% occlusion reconstruction}
\end{subfigure}
\begin{subfigure}[73\% occlusion reconstruction]{
\centering{\includegraphics[width = \textwidth]{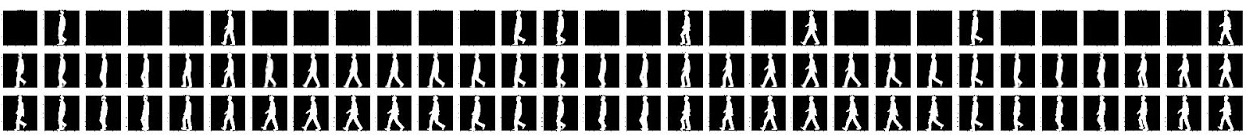}}
\label{rec_img:b}
}
%\caption{73\% occlusion reconstruction}
\end{subfigure}
\begin{subfigure}[73\% occlusion reconstruction]{
\centering{\includegraphics[width = \textwidth]{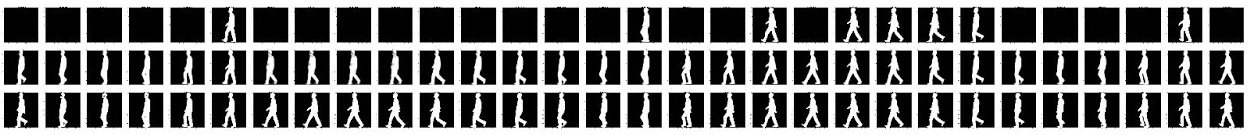}}
\label{rec_img:c}
}
%\caption{73\% occlusion reconstruction}
\end{subfigure}\label{rec_img}

\caption{Occlusion reconstruction results using BGaitR-NET on synthetically occluded CASIA-B data}
\label{rec_img}
\end{figure}

The BGaitR-NET model is trained on OU-ISIR\cite{Iwama2012Jun} and CASIA-B \cite{Yu06aframework} dataset with synthetic occlusions introduced in the dataset . The methodology adopted for training is briefly described in Section \ref{occlusionReconstruction}. The trained Encoder model and the Bi-LSTM model are deployed directly in our Proposed algorithm according to the schematic diagram in Figure \ref{fig:proposed_approach}. The high reconstruction quality of our BGaitR-NET model for even high degrees of occlusion can be visualized from Figure \ref{rec_img:a}, \ref{rec_img:b}, and \ref{rec_img:c}. In each sub-figure, the first row shows a gait cycle with missing frames (the percentage of missing frames in a gait cycle is given in the caption below each sub-figure) from the CASIA-B data, whereas the second and third row corresponds to the BGaitR-NET reconstructed output and the ground truth sequence (i.e., the sequence on which synthetic occlusion has been applied), respectively. Here, Figure \ref{rec_img:a} shows reconstruction of a sequence corrupted with 53\% occlusion, while Figures \ref{rec_img:b} and \ref{rec_img:c} respectively show gait cycle reconstruction with 73\% occlusion. For ease  of visualization, we have not shown all the frames present in the above three sequences. Still it can be seen from the figures that the frame reconstruction is indeed of very high quality. This is due to the incorporation of spatio-temporal information of a gait cycle while training the reconstruction model.

The Conditional Variational Auto-Encoder was trained on the OU-ISIR Large Population Dataset and the publicly available CASIA-B gait dataset. The OU-ISIR large Population Dataset contains 3254 subjects with a gallery viewpoint of 85 degrees (closest to the fronto-parallel view). We have used 2400 different subjects who were within the age limit of 15-75 to better fit the target subjects. Out of this, we used 2200 of them for the training set and the remaining 200 for the validation set. These frames were used as the input of the CVAE and the ground truth for the model. The training was done with Adam optimizer for 100 epochs with a learning rate of 0.01 and had converged on both training and validation sets to give a dice score of 0.972. For the Encoder Network, we used CASIA-B as the test set to get a dice score of 0.987. The mathematical calculation of a dice score($M_{dice}$) is given by \ref{dice_metrics}

\begin{equation} \label{dice_metrics}
    M_{dice} =  \frac{2F_i\hat{F_i}}{F_{i}^{2} + \hat{F_i^2}}
\end{equation}

Then the BiLSTM is trained on this latent vector obtained by running Encoder Model on the CASIA-B dataset. This dataset consists of 124 subjects with 6 walking sequences corresponding to each person. So a total of 744 gait sequences, was further broken down into individual sequences of 6 consecutive frames this summed up to 69560 sequences for the BiLSTM. The dataset was split into 65000 training sequences and 4560 validation sequences. We used an ADAM optimizer for 100 epochs with a learning rate of 0.01 for converging both the train and validation losses.

To quantify the effectiveness of the BGaitR-Net we test the reconstructed sequences with the accuracy obtained in gait recognition. For this, we employ a Convolutional Neural Network that takes the input of Gait Energy Image GEI\cite{Han2006Feb} computed from the whole sequence. This is trained with the subjects of the whole gallery. This trained model is used to predict the class of an unknown subject during the testing phase. This model has an output of softmax and has been trained with Multi-Class-Cross-Entropy loss on the given classes with Adam optimizer with a learning rate of 0.01. 
%With an occluded sequence \textbf{Figure, 4(a)} gave a reconstructed sequence of Figure 4(b). By Visual Inspection of the reconstructed output Figure, 4(b) to the Ground-Truth Figure 4(c) epitomizes the high efficiency of the Reconstruction capability of BGaitR-Net. 
Good quality of reconstruction have been seen to be consistent with the occluded sequences present in the TUM-IITKGP dataset as well. However, since this data set consists of real occlusion, and there is no ground truth sequence to compare the predicted sequence with the ground truth sequence, we have not presented these results in the paper.
%we shall not be able to visually inspect as there is no ground truth.

\subsection{Effectiveness of the Gait Reconstruction by gait recognition}\label{gaitRestrgaitRecog}

In our next experiment, we evaluate the GEI-Net-based gait recognition accuracy on the reconstructed sequences. 
For the CASIA-B data, we show the results for varying levels of synthetic occlusion between 0-90\% in Table \ref{casia_rank1}. Out of the six normal walking sequences present in the CASIA-B data (\textit{nm-01} to \textit{nm-06}), four gait sequences (namely, \textit{nm-01} to \textit{nm-04}) from each of the 124 subjects have been used for training the \textit{GEI-Net} model, and the remaining two (\textit{nm-05} and \textit{nm-06}) are used for testing after applying synthetic occlusion.

\begin{table}[]
  \centering
  \caption{Rank 1 gait recognition accuracy on unoccluded sequences of CASIA-B data as well as on occluded sequences for varying levels of synthetic occlusion on the CASIA-B data}
  \label{casia_rank1}
  \begin{tabular}{|c|c|}
    \hline
    Occlusion Degree (\%)& Rank 1 Accuracy (\%) on %Rank 1 Accuracy (\%) 
    \\
    %Degree (\%)& Sequences 
    & Synthetically Occluded Sequences\\
    \hline
    $\leq$10 &99.83 \\
    10 - 20 &99.53 \\
    20 - 30 &99.32 \\
    30 - 40 &97.16 \\
    40 - 50 &95.00 \\
    50 - 60 &93.21 \\
    60 - 70 &91.22 \\
    70 - 80 &76.65 \\
    80 - 90 &60.05 \\
    \hline
\end{tabular}
\end{table}

The Rank 1 Accuracy for classification is obtained from the GEI-Net and reported in Table \ref{casia_rank1}. Then this network is also used for TUM-IITKGP after fine-tune learning for the classes in the TUM-IITKGP dataset, The proposed algorithm remains the same with only minor adjustments done to the preprocess module that helps to parse the video input into cropped silhouettes for BGaitR-Net to function. This gave a cross-validation accuracy of 97.32\% on classifying the reconstructed sequence. It may be noted a better gait recognition accuracy can be obtained by tuning the network and use PEIs\cite{Roy2012Mar} over GEI\cite{Han2006Feb} for gait recognition. This may be considered as future scope for experimenting.

\subsection{Rank wise Accuracy of Gait recognition model}\label{rank5Gait}

\begin{figure}[]
\centerline{\includegraphics[scale=0.50]{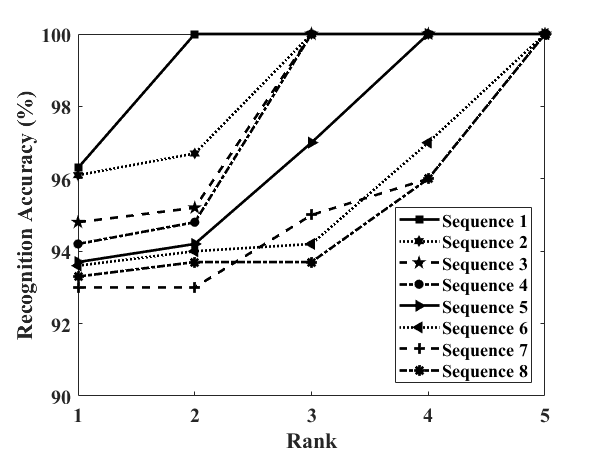}}
\caption{Rank-wise improvement in the accuracy of our algorithm on the eight occluded sequences present in the TUM-IITKGP dataset considering four normal walking sequences for training our GEInet model}
\label{img:cmc}
\end{figure}

The rank one accuracy (as computed in the previous experiment) is not always a reliable metric to evaluate the performance of an algorithm. Rather, the improvement in recognition accuracy with increment in rank by means of Cumulative Match Characteristic (CMC) curves provides important information regarding the effectiveness of the overall algorithm. Hence we perform experiments till Rank-5 for the TUM-IITKGP dataset. We have used the eight GEIs computed from the reconstructed sequence from the eight occluded sequences names, Sequence 1, Sequence 2, …, Sequence 8. The methodology remains the same, only now we take into account the top 5 predictions from the GEI-Net and plot them on the CMC curve as shown in Figure \ref{img:cmc}.

It can be seen the Rank 1 accuracy for recognition is greater than or equal to 93\% for all eight sequences. For Rank 2 and Rank 3 the minimum accuracy of all the sequences remains negligibly the same as the previous Rank. For Rank 4 all the sequences have an accuracy greater or equal to 96\%. And finally, for Rank 5 All the sequences are predicted accurately with 100\% accuracy.

\subsection{Robustness and Generalization of the Reconstruction model}\label{robustReconstruction}
\begin{figure}[]
\centerline{\includegraphics[scale=0.6]{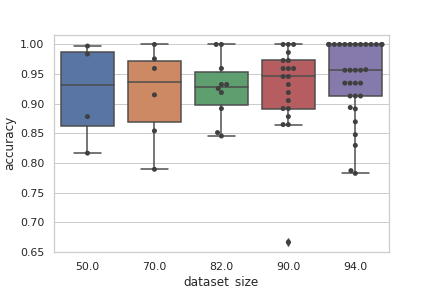}}\label{img:box_plot}
\caption{Recognition accuracy from Reconstruction on \% of the dataset}
\end{figure}

In our next experiment, we continue to test the robustness of our reconstruction model and show how much it generalizes across multiple data set of varying amount of the same image data set. For checking the robustness of our model we tested out using the  Stratified K fold Validation. This is we divide the dataset into $K$ equal parts randomly, then we select one of the parts and consider that as the test dataset while we train the model on the remaining $K - 1$ parts, This process is done for all the $K$ parts hence giving us $K$ different reading about the accuracy and performance of the model using Variance of all these accuracies. We have used 5 values for $K$ and presented the result in form of a box plot in Figure \ref{img:box_plot} to illustrate the robustness of the model. The 5 values of $K$ are $[2, 3, 5, 10, 16]$ for which each batch is trained on $[50\%, 70\%, 82\%, 90\%, 94\%]$ of the dataset respectively. For this experiment we have already trained Gait recognition GEI-Net model which is used for the final evaluation of all reconstruction models trained using the Stratisfied K Fold approach. So for individual Value of K there are K number of model trained on 100*(K-1)/K \% of the dataset and tested on 100/K \% of the dataset. This gives use K values of accuracies for each K used. These accuracies are then plotted using a box plot which plots the relation between the Mean and variance of the accuracies achieved in form of Quartile Deviation.

It is clear from the curve that there is a very small amount of increase in the average accuracy from 93\% by training on 50\% of the dataset and increases steadily to a average of 96\% by having trained on 94\% of the dataset. We can also see a decrease in variance of the given experiments. There is no perfect trend in the Variance because the model may have randomly made divisions of the dataset were a certain fold may have a very different distribution compared to the training set, But it is also visible there are only a few of those which are making the variance high. The Upper Quartile is increasing in accuracy by increasing the size of the training set.

\subsection{Comparative analysis of our Proposed approach}\label{comparative}
Further to get a baseline for the currently proposed algorithm we present a comparative analysis of our work with some popular gait recognition techniques ( with and without occlusion handling methods).  To date, only a few occlusion handling techniques have been developed for gait recognition, and the work of \cite{Babaee2019Apr}, and \cite{Babaee2018Oct} are few important among these that deal with occlusion reconstruction and are used in the comparative study. The work in \cite{Chen2009Aug}, discusses an effective method to extract gait features from self-occluded sequences this has also been used in the comparative study. However, the occlusion handling method described in \cite{Chattopadhyay2015Oct} could not be used in this study since it focuses only on frontal gait recognition using the depth and skeleton streams provided by Kinect, whereas our present work considers only binary silhouette sequence for gait cycle reconstruction and feature extraction. To verify the importance of developing a suitable occlusion reconstruction mechanism, we also compare the results of our work with that of some traditional gait recognition algorithms using the same set of test occluded sequences from the TUM-IITKGP data. A major contribution with such an approach is the work described in \cite{Han2006Feb}, which uses a simple feature aggregation technique to compute gait features. This work serves as the baseline of Computer Vision-based gait recognition and performs with very high accuracy on clean unoccluded sequences. We also compare our work with two other non-deep Learning-based approaches, namely \cite{Roy2012Mar} and \cite{Zhang2010Jul}, and two recent Deep Learning-based gait recognition approaches, namely, \cite{Alotaibi2017Nov} and \cite{Shiraga2016Jun}.

\begin{table}[]
  \centering
  \caption{Comparative analysis of the proposed work with existing approaches on the TUM-IITKGP dataset}
  
  \begin{tabular}{|c|c|c|c|}
    \hline
    \ttfamily Category & \ttfamily Method & \ttfamily Average Acc. & \ttfamily Time (secs) \\
    \hline
    \multirow{4}{4cm}{Method with Occlusion Handling Mechanism} & \multicolumn{1}{c|}{Proposed BGaitR-Net} &  \multicolumn{1}{c|}{97.32 \%} &  \multicolumn{1}{c|}{0.63} \\
     &  \multicolumn{1}{c|}{\cite{Babaee2018Oct}} &  \multicolumn{1}{c|}{78.92 \%} &  \multicolumn{1}{c|}{0.22} \\
     &  \multicolumn{1}{c|}{\cite{Babaee2019Apr}} &  \multicolumn{1}{c|}{80.00 \%} &  \multicolumn{1}{c|}{0.35} \\
     &  \multicolumn{1}{c|}{\cite{Chen2009Aug}} &  \multicolumn{1}{c|}{77.65 \%} &  \multicolumn{1}{c|}{0.10}  \\
    \hline
    \multirow{5}{4cm}{Methods without Occlusion Handling Mechanism} & \multicolumn{1}{c|}{\cite{Han2006Feb}} & \multicolumn{1}{c|}{65.71 \%} &  \multicolumn{1}{c|}{0.09} \\
     &  \multicolumn{1}{c|}{\cite{Roy2012Mar}} &  \multicolumn{1}{c|}{70.23 \%} &  \multicolumn{1}{c|}{0.25} \\
     &  \multicolumn{1}{c|}{\cite{Zhang2010Jul}} &  \multicolumn{1}{c|}{73.54 \%} &  \multicolumn{1}{c|}{0.10} \\
     &  \multicolumn{1}{c|}{\cite{Alotaibi2017Nov}} &  \multicolumn{1}{c|}{76.42 \%} &  \multicolumn{1}{c|}{2.59} \\
     &  \multicolumn{1}{c|}{\cite{Shiraga2016Jun}} &  \multicolumn{1}{c|}{76.79 \%} &  \multicolumn{1}{c|}{3.09}  \\
    \hline
\end{tabular}
\end{table}

Results are reported in Table 2 in terms of Average Rank 1 accuracy. Using the same evaluation method explained at the start of [Section 4] for Rank 1 accuracy of TUM-IITKGP dataset. It is seen from the table that the proposed approach outperforms each of the other gait recognition approaches in terms of recognition accuracy by a substantial margin.
Among the gait recognition methods with occlusion handling capability, the work in \cite{Babaee2019Apr} and \cite{Babaee2018Oct} attempts to predict the complete GEI of a subject from the incomplete GEIs corresponding to the available clean/occluded frames in a sequence by employing a deep neural network. The effectiveness of these approaches depends on the quality and the number of initial clean frames available for a reasonably good initial GEI feature, which can be further refined using the deep network. However, due to the usage of averaged silhouette information and ignoring the detailed frame-level information during the reconstruction process, the quality of the reconstructed frames gets degraded to a certain extent. As seen in the table, these methods still perform with reasonably good Rank 1 accuracy of 80.00\% and 78.56\%, respectively. In the proposed work, we reconstruct occluded frames using a BiLSTM model while considering all the different frames of the sequence instead of a single averaged GEI. Our approach also doesn’t make any assumption about the gait cycle and produces temporally consistent frames. The work in \cite{Chen2009Aug} is appropriate in situations with small degrees of self-occlusion. It does not reconstruct occluded frames, rather computes the difference between adjacent frames to reduce the effect of self-occlusion, and is not appropriate for application in situations where the whole silhouette information in a few frames may be missing. Among the recent deep learning-based gait recognition techniques used in the present study, \cite{Alotaibi2017Nov}, \cite{Shiraga2016Jun} have comparable performance, and these also improve upon the recognition accuracy given by the appearance-based gait features, i.e., \cite{Han2006Feb}, \cite{Roy2012Mar}, and \cite{Zhang2010Jul}. The proposed occlusion reconstruction using BGaitR-Net is significantly accurate and outperforms each of the other approaches used in the comparative study in terms of average accuracy.
The results also demonstrated were tested out on modest hardware as described and we observed the average time for running the proposed algorithm was only 0.12 seconds.

\section{Conclusions and Future Work}

In this work, we have proposed an algorithm to reconstruct a corrupted gait sequence, that could be integrated with any gait recognition system to improve performance in a real-life scenario. We have used Artificial Neural Networks for Encoding and reconstructing the frame, which takes advantage of the high accuracy and generalization capability of deep neural networks. The encoder-decoder model used is a Conditional-Variational AutoEncoder which uses the key pose of the image as the conditional vector to gives a compact encoded vector for the image that could be used in place of the traditional PCA projection in the EigenSpace. On the other hand, BiLSTM filters the encoded sequence of frames to reconstruct any form of occlusions. With appearance-based recognition, we have also used model-based features like Key poses which gave us better reconstruction in regions where there was a high degree of occlusion. To the best of our knowledge, ours is the first work that employs deep neural networks to reconstruct a corrupted gait sequence with both Appearance and model-based features fused together to produce Temporally consistent sequences. In the future, work can be extended by fitting complex kinematics models on the gait cycle whose parameters can be fused with the Silhouette features to obtain better encoding in the latent space leading to better results while filtering the whole gait cycle with BiLSTMs. The proposed occlusion reconstruction algorithm can potentially be applied in video surveillance applications like gait recognition and also other applications like person re-identification, activity recognition, etc

\section{Acknowledgments}

The authors would like to thank NVIDIA for supporting their research with a Titan XP GPU.

 \bibliographystyle{elsarticle-num} 
 \bibliography{cas-refs}

\begin{thebibliography}{10}
\expandafter\ifx\csname url\endcsname\relax
  \def\url#1{\texttt{#1}}\fi
\expandafter\ifx\csname urlprefix\endcsname\relax\def\urlprefix{URL }\fi
\expandafter\ifx\csname href\endcsname\relax
  \def\href#1#2{#2} \def\path#1{#1}\fi

\bibitem{Chattopadhyay2015Oct}
P.~Chattopadhyay, S.~Sural, J.~Mukherjee, {Frontal Gait Recognition from
  Occluded Scenes}, Pattern Recognition Letters 63 (2015) 9--15.

\bibitem{Babaee2019Apr}
M.~Babaee, L.~Li, G.~Rigoll, {Person Identification from Partial Gait Cycle
  Using Fully Convolutional Neural Networks}, Neurocomputing 338 (2019)
  116--125.

\bibitem{Roy2011Nov}
{Roy, Aditi and Sural, Shamik and Mukherjee, Jayanta and Rigoll, Gerhard},
  {Occlusion Detection and Gait Silhouette Reconstruction from Degraded
  Scenes}, Signal, Image, and Video Processing 5~(4) (2011) 415--430.

\bibitem{Zheng2011Sep}
S.~Zheng, J.~Zhang, K.~Huang, R.~He, T.~Tan, {Robust View Transformation Model
  for Gait Recognition}, in: Proceedings of the $18^{th}$ IEEE International
  Conference on Image Processing, 2011, pp. 2073--2076.

\bibitem{Yu06aframework}
S.~Yu, D.~Tan, T.~Tan, {A Framework for Evaluating the Effect of View Angle,
  Clothing and Carrying Condition on Gait Recognition}, in: Proceedings of the
  International Conference on Pattern Recognition, 2006, pp. 441--444.

\bibitem{Iwama2012Jun}
H.~Iwama, M.~Okumura, Y.~Makihara, Y.~Yagi, {The OU-ISIR Gait Database
  Comprising the Large Population Dataset and Performance Evaluation of Gait
  Recognition}, IEEE Transactions on Information Forensics and Security 7~(5)
  (2012) 1511--1521.

\bibitem{Hofmann2011GaitRI}
M.~Hofmann, S.~Sural, G.~Rigoll, {Gait Recognition in the Presence of
  Occlusion: A New Dataset and Baseline Algorithms}, in: Proceedings of the
  $19^{th}$ International Conference in Central Europe on Computer Graphics,
  Visualization and Computer Vision, 2011, pp. 99--104.

\bibitem{Han2006Feb}
J.~Han, B.~Bhanu, {Individual Recognition Using Gait Energy Image}, IEEE
  Transactions on Pattern Analysis and Machine Intelligence 28~(2) (2006)
  316--322.

\bibitem{Roy2012Mar}
A.~Roy, S.~Sural, J.~Mukherjee, {Gait Recognition Using Pose Kinematics and
  Pose Energy Image}, Signal Processing 92~(3) (2012).

\bibitem{Chattopadhyay2014Jan}
P.~Chattopadhyay, A.~Roy, S.~Sural, J.~Mukhopadhyay, {Pose Depth Volume
  Extraction from RGB-D Streams for Frontal Gait Recognition}, Journal of
  Visual Communication and Image Representation 25~(1) (2014) 53--63.

\bibitem{Zhang2010Jul}
E.~Zhang, Y.~Zhao, W.~Xiong, {Active Energy Image Plus 2DLPP for Gait
  Recognition}, Signal Processing 90~(7) (2010) 2295--2302.

\bibitem{Xu2007Oct}
D.~Xu, S.~Yan, D.~Tao, S.~Lin, H.-J. Zhang, {Marginal Fisher Analysis and Its
  Variants for Human Gait Recognition and Content- Based Image Retrieval}, IEEE
  Trans. Image Process. 16~(11) (2007) 2811--2821.

\bibitem{Collins2002May}
R.~T. Collins, R.~Gross, J.~Shi, {Silhouette-Based Human Identification from
  Body Shape and Gait}, in: {Proceedings of $5^{th}$ IEEE International
  Conference on Automatic Face Gesture Recognition}, 2002, pp. 366--371.

\bibitem{Ariyanto2011Oct}
G.~Ariyanto, M.~S. Nixon, {Model-Based 3D Gait Biometrics}, in: Proceedings of
  the International Joint Conference on Biometrics, 2011, pp. 1--7.

\bibitem{Cunado2005Jun}
D.~Cunado, M.~S. Nixon, J.~N. Carter, {Using Gait as a Biometric, via
  Phase-Weighted Magnitude Spectra}, in: {Proceedings of the Audio-and
  Video-Based Biometric Person Authentication}, 2005, pp. 93--102.

\bibitem{Bobick2001Dec}
A.~F. Bobick, A.~Y. Johnson, {Gait Recognition Using Static, Activity-Specific
  Parameters}, in: {Proceedings of the IEEE Computer Society Conference on
  Computer Vision and Pattern Recognition}, 2001, pp. I--423--I--430.

\bibitem{Cunado2003Apr}
D.~Cunado, M.~S. Nixon, J.~N. Carter, {Automatic Extraction and Description of
  Human Gait Models for Recognition Purposes}, Computer Vision and Image
  Understanding 90~(1) (2003) 1--41.

\bibitem{Lee2002May}
L.~Lee, W.~E.~L. Grimson, {Gait Appearance for Recognition}, in: Proceedings of
  the International Workshop on Biometric Authentication, 2002, pp. 143--154.

\bibitem{Hu2018}
H.~Hu, Y.~Li, Z.~Zhu, G.~Zhou, {CNNAuth: Continuous Authentication via
  Two-Stream Convolutional Neural Networks}, in: Proceedings of the IEEE
  International Conference on Networking, Architecture and Storage, 2018, pp.
  1--9.

\bibitem{YantaoLiJul2020}
Y.~Li, H.~Hu, Z.~Zhu, G.~Zhou, {SCANet: Sensor-Based Continuous Authentication
  With Two-Stream Convolutional Neural Networks}, ACM Transactions on Sensor
  Networks 16~(3) (2020) 1--27.

\bibitem{Qin2021}
H.~Qin, M.~A. El-Yacoubi, Y.~Li, C.~Liu, {Multi-Scale and Multi-Direction GAN
  for CNN-Based Single Palm-Vein Identification}, IEEE Transactions on
  Information Forensics and Security 16 (2021) 2652--2666.

\bibitem{Takemura2017Oct}
N.~Takemura, Y.~Makihara, D.~Muramatsu, T.~Echigo, Y.~Yagi, {On Input/Output
  Architectures for Convolutional Neural Network-Based Cross-View Gait
  Recognition}, IEEE Transactions on Circuits and Systems for Video Technology
  29~(9) (2017) 2708--2719.

\bibitem{Shiraga2016Jun}
K.~Shiraga, Y.~Makihara, D.~Muramatsu, T.~Echigo, Y.~Yagi, {GEINet:
  View-Invariant Gait Recognition Using a Convolutional Neural Network}, in:
  {Proceedings of the International Conference on Biometrics}, 2016, pp. 1--8.

\bibitem{Alotaibi2017Nov}
M.~Alotaibi, A.~Mahmood, {Improved Gait Recognition Based on Specialized Deep
  Convolutional Neural Network}, Computer Vision and Image Understanding 164
  (2017) 103--110.

\bibitem{Sivapalan2011Oct}
S.~Sivapalan, D.~Chen, S.~Denman, S.~Sridharan, C.~Fookes, {Gait Energy Volumes
  and Frontal Gait Recognition Using Depth Images}, in: {Proceedings of the
  International Joint Conference on Biometrics}, 2011, pp. 1--6.

\bibitem{Battistone2019Sep}
F.~Battistone, A.~Petrosino, {TGLSTM: A Time Based Graph Deep Learning Approach
  to Gait Recognition}, Pattern Recognition Letters 126 (2019) 132--138.

\bibitem{Isa2010Dec}
W.~N.~M. Isa, M.~J. Alam, C.~Eswaran, {Gait Recognition Using Occluded Data},
  in: {Proceedings of the Asia Pacific Conference on Circuits and Systems},
  2010, pp. 344--347.

\bibitem{Lee2009Dec}
T.~K.~M. Lee, M.~Belkhatir, S.~Sanei, {Coping with Full Occlusion in
  Fronto-Normal Gait by Using Missing Data Theory}, in: {Proceedings of the
  $7^{th}$ International Conference on Information, Communications and Signal
  Processing}, 2009, pp. 1--5.

\bibitem{Aly2014Dec}
S.~Aly, {Partially Occluded Pedestrian Classification Using Histogram of
  Oriented Gradients and Local Weighted Linear Kernel Support Vector Machine},
  IET Computer Vision 8~(6) (2014) 620--628.

\bibitem{Zhang2010Aug}
J.~Zhang, H.~Sun, W.~Guan, J.~Wang, Y.~Xie, B.~Shang, {Robust Human Tracking
  Algorithm Applied for Occlusion Handling}, in: {Proceedings of the $5^{th}$
  International Conference on Frontier of Computer Science and Technology},
  2010, pp. 546--551.

\bibitem{Andriluka2008Jun}
M.~Andriluka, S.~Roth, B.~Schiele, {People-Tracking-by-Detection and
  People-Detection-by-Tracking}, Proceedings of the IEEE Conference on Computer
  Vision and Pattern Recognition (2008) 1--8.

\bibitem{deLeon2002Aug}
R.~D. de~Le{\ifmmode\acute{o}\else\'{o}\fi}n, L.~E. Sucar, {Continuous Activity
  Recognition With Missing Data}, in: {Proceedings of the International
  Conference on Pattern Recognition}, 2002, pp. 439--442.

\bibitem{Weinland2010Sep}
D.~Weinland, M.~{\ifmmode\ddot{O}\else\"{O}\fi}zuysal, P.~Fua, {Making Action
  Recognition Robust to Occlusions and Viewpoint Changes}, in: European
  Conference on Computer Vision, 2010, pp. 635--648.

\bibitem{Schuster1997Dec}
M.~Schuster, K.~K. Paliwal, {Bidirectional Recurrent Neural Networks}, IEEE
  Transactions on Signal Processing 45~(11) (1997) 2673--2681.

\bibitem{Pagnoni2018Dec}
A.~Pagnoni, K.~Liu, S.~Li, {Conditional Variational Autoencoder for Neural
  Machine Translation}, arXiv preprint arXiv:1812.04405 (2018).

\bibitem{Kingma2013}
D.~P. Kingma, M.~Welling, {Stochastic Gradient VB and the Variational
  Auto-Encoder}, in: Proceedings of the $2^{nd}$ International Conference on
  Learning Representations, Vol.~19, 2014, p. 121.

\bibitem{shiraga2016geinet}
K.~Shiraga, Y.~Makihara, D.~Muramatsu, T.~Echigo, Y.~Yagi, {GEINet:
  View-Invariant Gait Recognition Using a Convolutional Neural Network}, in:
  Proceedings of the International Conference on Biometrics, IEEE, 2016, pp.
  1--8.

\bibitem{Babaee2018Oct}
M.~Babaee, L.~Li, G.~Rigoll, {Gait Recognition from Incomplete Gait Cycle}, in:
  Prceedings of the $25^{th}$ IEEE International Conference on Image
  Processing, 2018, pp. 768--772.

\bibitem{Chen2009Aug}
C.~Chen, J.~Liang, H.~Zhao, H.~Hu, J.~Tian, {Frame Difference Energy Image for
  Gait Recognition with Incomplete Silhouettes}, Pattern Recognition Letters
  30~(11) (2009) 977--984.

\end{thebibliography}





% \end{thebibliography}
\end{document}